\documentclass[
]{ceurart}

\usepackage{listings}
\usepackage{afterpage}
\usepackage{multicol}
\usepackage{lipsum}
\usepackage{caption}

\usepackage[dvipsnames]{xcolor}
\definecolor{GPT4}{HTML}{33BBEE}
\definecolor{PF}{HTML}{4477AA}
\colorlet{Mycolor1}{green!10!orange}
\definecolor{GPT3}{HTML}{EE7733}

\usepackage{subcaption}  
\usepackage{float}

\begin{document}

\copyrightyear{2024}
\copyrightclause{Copyright for this paper by its authors.
  Use permitted under Creative Commons License Attribution 4.0
  International (CC BY 4.0).}

\conference{CHR 2024: Computational Humanities Research Conference, December 4–6, 2024, Aarhus, Denmark}

\title{Does ChatGPT Have a Poetic Style?}

\author[1]{Melanie Walsh}[%
orcid=0000-0003-4558-3310,
email=melwalsh@uw.edu,
url=,
]
\address[1]{Information School, University of Washington
  }
\author[2]{Anna Preus}[%
orcid=,
email=apreus@uw.edu,
url=,
]
\address[2]{English Department, University of Washington}
\author[1]{Elizabeth Gronski}[%
orcid=,
email=,
url=,
]

\cortext[1]{Corresponding author.}

\begin{abstract}
Generating poetry has become a popular application of LLMs, perhaps especially of  OpenAI's widely-used chatbot ChatGPT. What kind of poet is ChatGPT?
Does ChatGPT have its own poetic style?
Can it successfully produce poems in \textit{different} styles? 
To answer these questions, we prompt the GPT-3.5 and GPT-4 models to generate English-language poems in 24 different poetic forms and styles, about 40 different subjects,  and in response to 3 different writing prompt templates. We then analyze the resulting 5.7k poems, comparing them to a sample of 3.7k poems from the Poetry Foundation and the Academy of American Poets. 
We find that the GPT models, especially GPT-4, can successfully produce poems in a range of both common and uncommon English-language forms in superficial yet noteworthy ways, such as by producing poems of appropriate lengths for sonnets (14 lines), villanelles (19 lines), and sestinas (39 lines).
But the GPT models also exhibit their own distinct stylistic tendencies, both within and outside of these specific forms.
Our results show that GPT poetry is much more constrained and uniform than human poetry, showing a strong penchant for rhyme, quatrains (4-line stanzas), iambic meter, first-person plural perspectives (we, us, our), and specific vocabulary like ``heart,'' ``embrace,'' ``echo,'' and ``whisper.''


\end{abstract}


\maketitle

\section{Introduction}

\begin{figure}[h]
    \centering
    \includegraphics[width=.58\linewidth]{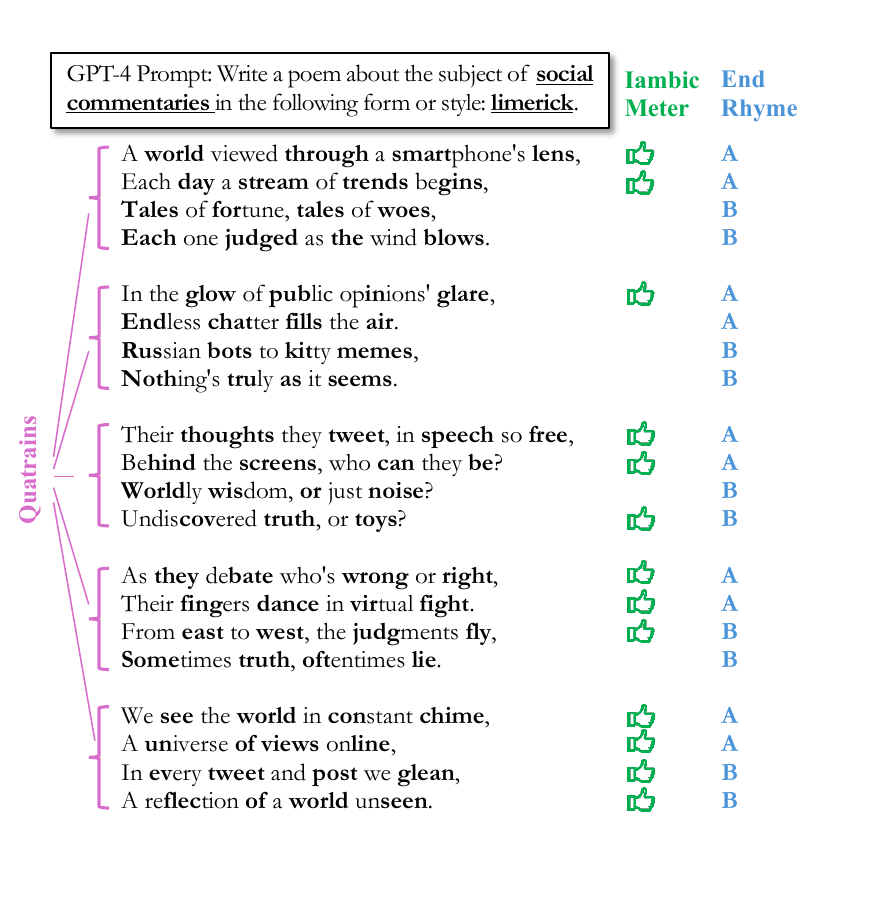}
    \caption{\textbf{``Write a poem about the subject of \textit{social commentaries} in the following form or style: \textit{limerick}.''} An example poetry generation prompt and response by GPT-4, representing common tendencies of the model. While GPT-4 presents a comedic, topical take on a social commentary, it does not produce a typical \textit{limerick} (usually 5 lines, anapestic meter, AABBA rhyme) but instead produces five quatrains with mostly iambic meter and AABB rhyme, what we suggest is its ``default'' mode.}
    \label{fig:intro}
\end{figure}

Generating poetry has become a popular application of LLMs, perhaps especially of OpenAI's widely-used ChatGPT. The chatbot's ability to produce poetry was frequently mentioned in early journalistic reports \cite{zahn_what_2022}, and the question of whether it produced ``good'' or ``bad'' poetry quickly became the subject of hot debate \cite{hunter_what_2023, clarke_chatgpt_2023}. 
While these debates point to the significance of LLM-generated poetry as a topic, in this paper we are less interested in subjectively evaluating the quality of ChatGPT-generated poems and more interested in quantifying and describing key patterns in the poetry produced.
What kind of poet is ChatGPT? 
Does ChatGPT have its own poetic style? 
Can it successfully produce poems in different forms and styles, like sonnets (14 lines with regular rhyme and iambic pentameter) or villanelles (19 lines with alternating first and third lines)?

In this paper, we present initial results from a larger study of LLM-generated poetry.
We prompt the GPT-3.5 and GPT-4 models to generate English-language poems in 24 different poetic forms or styles, about 40 different subjects, and in response to 3 different templates. We then analyze the resulting 5.7k poems, comparing them to a sample of 3.7k poems from the Poetry Foundation and the Academy of American Poets. 
We find that the GPT models, especially GPT-4, can successfully produce poems in a range of both common and uncommon English-language forms in superficial yet noteworthy ways, such as by producing poems of appropriate lengths for sonnets (14 lines), villanelles (19 lines), and sestinas (39 lines).
But we find that the GPT models also exhibit their own distinct stylistic tendencies, both within and outside of these specific forms.

Our results show that the poetry produced by GPT-3.5 and GPT-4 is much more constrained and uniform than human poetry. 
Unless otherwise prompted (and sometimes when otherwise prompted), both GPT models have a tendency to produce rhymed lines in something like iambic meter---a regular pattern of unstressed and stressed syllables that characterizes the majority of English-language verse before the 20th century \citep{tarlinskaja_meter_1987}. 
Both models also have a tendency to organize poetic lines into quatrains (4-line stanzas). 
And they display other distinct signatures, such as a curiously dominant first-person plural perspective and a penchant for words like ``heart,'' ``embrace,'' ``echoes,'' and ``whispers.''
We release our code,\footnote{\href{https://github.com/melaniewalsh/chatgpt\_poetry}{https://github.com/melaniewalsh/chatgpt\_poetry}} and we plan to share our ChatGPT poetry corpus at a later stage of the project, with the aim of encouraging further analysis through computational and/or more traditional literary studies approaches.



\section{Related Work}

The history of computational poetry generation dates back to at least the mid-20th century \cite{joyce_poetry_1975}. Poets, researchers, and hobbyists have experimented with a range of technical approaches  \cite{goncalo_oliveira_automatic_2009}, from rule-based systems \cite{manurung_chart_1999, gervas_expert_2001}, to Markov chains \cite{astigarraga_markov_2017}, to most recently neural networks and LLMs \cite{wang_chinese_2016, yi_automatic_2018, lau_deep-speare_2018, jhamtani_learning_2019, agarwal_acrostic_2020, chakrabarty_help_2022, ormazabal_poelm_2022, belouadi_bygpt5_2023}. 

While computational poetry has been an active area of inquiry for many decades \cite{manurung_towards_2000, tsan_automatic_2008, goncalo_oliveira_tra--lyrics_2007, das_poetic_2014, parrish_articulations_2018, melanie-becquet_exploring_2024}, it has arguably remained the purview of specialists until recently. But since the release of ChatGPT in November 2022, hundreds of millions of people have used and experimented with LLMs, opening up computational poetry generation to a broader public. This development has prompted emerging research on the narrative and literary qualities of LLM-generated texts \cite{tian_are_2024} and specifically LLM-generated poetry \cite{martinHistoricalProsodyMechanical2024, chen_evaluating_2024}.

Training data and memorization are key considerations for LLM-generated poetry from ethical, legal, and technical perspectives. Models' ability to produce poetry is intimately tied to their training data, which partly consists of literary works by both living and dead writers.
Much popular and scholarly attention related to LLMs and literature has focused---rightfully---on the ethics and legality of such literary training data \cite{tanRethinkingLiteraryPlagiarism2024, kulkarni_ethical_2022, veltman_ai_2024, vynck_ai_2023}. We believe these concerns are vitally important to examinations of ChatGPT's style, which is built from the words of other writers. We also think that it is valuable to ask questions about LLMs' poetic capacities because it can help inform debates about LLMs and creativity while also advancing our understanding of how poetry is being used and propagated in the contemporary world.

In their work on poetry memorization in ChatGPT specifically, \citet{dsouza_chatbot_2023} show that the most likely factor for a poem's memorization by the model was its inclusion in the 1983 \textit{Norton Anthology of Literature}. This finding suggests that canonical poetry is disproportionately represented in the GPT models, which could influence the kind of poetry they produce. In a similar vein, we show in prior work \citep{walsh_sonnet_2024} that 41\% of a curated sample of poems from the Poetry Foundation and the Academy of American Poets (which we also use as a comparison corpus in this study) are likely memorized by GPT-4. Our previous analysis suggests that this memorization may enhance the models' ability to classify the form of the poems, but the results are not conclusive. More work is needed to evaluate the impact that memorization may have on poetry generation.

\section{Data + Methods}

\subsection{Human Poetry Corpus}\label{human}

To provide a baseline comparison for our ChatGPT-generated poetry and to guide our prompting, we curate a dataset of poems, styles, and subjects from the Poetry Foundation and the Academy of American Poets. Both organizations are well-respected poetry institutions with websites that host tens of thousands of poems spanning hundreds of years, and many of the poems are tagged by style and subject on the websites. 

\begin{table*}
    \centering
    \scriptsize
        \caption{The distribution of poems by form and source.}

    \begin{tabular}{@{}m{1.6cm}|m{2.3cm}m{1.6cm}m{1.6cm}@{}}
    \toprule
    \textbf{Poetic Form x Source} & \textbf{Poetry \newline Foundation \&  \newline Academy of \newline American \newline Poets} & \textbf{GPT-3.5} & \textbf{GPT-4} \\
    \midrule
    \multicolumn{4}{@{}l}{\textbf{\textit{Fixed Forms}}} \\
    Ballad & 110 & 120 & 120 \\
    Ghazal & 40 & 120 & 120 \\
    Haiku & 50 & 120 & 120 \\
    Limerick & 7 & 120 & 120 \\
    Pantoum & 25 & 120 & 120 \\
    Sestina & 41 & 120 & 120 \\
    Sonnet & 856 & 120 & 120 \\
    Villanelle & 63 & 120 & 120 \\
    \midrule
    \multicolumn{4}{@{}l}{\textbf{\textit{Formal Elements}}} \\
    \textit{Meters} \\
    Blank Verse & 209 & 120 & 120 \\
    Free Verse & 387 & 120 & 120 \\
    Common Measure & 112 & 120 & 120 \\
    \textit{Stanza Forms} \\
    Couplet & 398 & 120 & 120 \\
    Quatrain & 89 & 120 & 120 \\
    Tercet & 94 & 120 & 120 \\
    \midrule
    \multicolumn{4}{@{}l}{\textbf{\textit{Unfixed Forms}}} \\
    Ars Poetica & 94 & 120 & 120 \\
    Aubade & 16 & 120 & 120 \\
    Concrete \newline Poetry & 24 & 120 & 120 \\
    Dramatic Monologue & 191 & 120 & 120 \\
    Ekphrasis & 145 & 120 & 120 \\
    Elegy & 254 & 120 & 120 \\
    Ode & 119 & 120 & 120 \\
    Pastoral & 75 & 120 & 120 \\
    Prose Poem & 475 & 120 & 120 \\
    ``A Poem'' & - & 120 & 120 \\
    \midrule
    \textbf{Total} & \textbf{3,874 \tiny{poem/form pairs}} & \textbf{2,880 \tiny{poems}} & \textbf{2,880 \tiny{poems}} \\
    \bottomrule
    \end{tabular}
    \label{table:poem-distribution}

\end{table*}

We scrape up to 400 poems from these two sources for 23 different poetic forms or styles, which we also use as prompts for our ChatGPT-generated poetry corpus. Following prior work \citep{walsh_sonnet_2024}, we select poems in the following categories: \textsc{fixed forms}, \textsc{unfixed forms}, and \textsc{formal elements} (which consists of both meters and stanza forms) (see Table \ref{table:poem-distribution}). 
In total, the sample includes 3,874 poem/style pairs, or 3,692 unique poems.

We manually remove prefatory text---such as dedications, dates, epigraphs, or other contextual information---from human-authored poems with traditionally fixed lengths (e.g., sonnets, villanelles, sestinas) if the poem is within 10 lines of the conventional length. We do not remove prefatory material from other poems; however, based on our qualitative analysis and review, we do not believe prefatory material is extensive in most other poems or significantly impacts results.

While the Poetry Foundation and the Academy of American Poets are among the largest tagged poetry collections available, they are also defined by various kinds of bias that are important to note. They both focus on English-language poetry, and the Academy of American Poets focuses especially on American poetry. Not all of the poems in their collections are tagged, and it is unclear why some poems are tagged and others are not. Neither site hosts a representative collection of poems, in terms of poets' gender, race, sexuality, and time period (it is also difficult to know what a representative collection would be). They also over-represent prestigious and canonical poetry, which may be of particular note in comparison with ChatGPT since the model may be trained on (and perhaps even encouraged to produce) more popular, commercial, and colloquial poetry.


\subsection{ChatGPT-Generated Poetry Corpus}

\begin{table*}
\caption{Subjects, styles, and writing prompt templates for the GPT-generated poetry corpus.}
  \begin{tabular}{p{0.3\linewidth} p{0.65\linewidth}}
    \toprule
    \textbf{Category} &  \\
    \midrule
    Subjects & \textbf{General:} activities, arts \& sciences, living, love, mythology \& folklore, nature, religion, relationships, social commentaries \newline
    \textbf{Occasions:} anniversary, birth, birthdays, engagement, farewells \& good luck, funerals, get well \& recovery, graduation, gratitude \& apologies, toasts \& celebrations, weddings \newline
    \textbf{Holidays:} cinco de mayo, christmas, easter, father's day, halloween, hanukkah, independence day, kwanzaa, memorial day, mother's day, new year, passover, ramadan, rosh hashanah, september 11th, st. patrick's day, thanksgiving, valentine's day, yom kippur \\
    \midrule
    Styles & \textbf{Fixed:} limerick, pantoum, ghazal, ballad, villanelle, sonnet, sestina, haiku \newline
    \textbf{Unfixed:} epic, monologue, ars poetica, aubade, pastoral, ode, elegy, visual poetry, ekphrasis, prose poem \newline
    \textbf{Formal Elements:} \textit{meters:} common measure, blank verse, free verse \newline
    \textit{stanza forms:} quatrain, tercet, couplet \\
    \midrule
    Prompt Templates &
    \textbf{General:} Write a poem about the subject of X in the following form or style: Y. \newline 
    \textbf{Figurative:} Write a poem about the subject of X in the following form or style: Y. Do not use the actual word(s) X or Y in the poem.\newline 
     \textbf{Specific:} Write a poem about the subject of X in the following form or style: Y. Make the poem about something specific.\\
    \bottomrule
  \end{tabular}
  \label{tab:subjects_styles_prompts}
\end{table*}

To create our ChatGPT poetry corpus, we prompt \textbf{GPT-3.5 Turbo} and \textbf{GPT-4} \cite{openai_gpt-4_2024} to generate poems in response to 3 different \textit{writing prompt templates}, in 24 different \textit{styles/forms}, and about 40 different \textit{subjects}. The styles and subjects are selected from the tagging schema on the Poetry Foundation's website. We use zero-shot prompts (i.e., prompts that do not provide desired example outputs) because we are interested in testing the model's ``out-of-the-box'' capabilities in a mostly unmediated form. 

We select styles and subjects from the Poetry Foundation because they offer an extensive and diverse poetic taxonomy that is developed by an authoritative external source and that is reflective of one of the largest existing collections of human poetry. For our ``subjects,'' we select the 40 broadest level ``topics'' from the Poetry Foundation's tagging schema, which include the subcategories ``subjects,'' ``occasions,'' and ``holidays'' (see Table \ref{tab:subjects_styles_prompts}). For our ``styles,'' we select the 23 styles and poetic forms described in Section \ref{human}. We add the style of ``a poem'' because we are interested in the models' responses to the generic idea of a poem without a specified form. These combinations result in 2,880 generated poems per model, with 120 poems per style (per model) and 72 poems per subject (per model) (see Table \ref{table:poem-distribution}).

 
We model the construction of our 3 \textit{writing prompt templates} on popular approaches demonstrated on social media, in journalistic articles, and by LLM companies \cite{jim_al-khalili_jimalkhalili_i_2023, hunter_what_2023, microsoft_how_2023}:
\begin{enumerate}
    \item \textbf{General:} Write a poem about the subject of X in the following form or style: Y.\\
    \item \textbf{Figurative:} Write a poem about the subject of X in the following form or style: Y. Do not use the actual word(s) X or Y in the poem.\\
     \item \textbf{Specific:} Write a poem about the subject of X in the following form or style: Y. Make the poem about something specific.\\
\end{enumerate}
We include our ``figurative`` and ``specific`` templates after observing the models' tendency to repeat the words in the prompts and to be vague. These templates push the model to create more diverse outputs.
We believe that prompting significantly impacts the kind of poetry that the GPT models produce, and we reflect on this more in Section \ref{discussion}.

\section{Results}

\subsection{Poetic Length \& Structure} \label{length}

\begin{figure}[htbp!]
    \centering
    \includegraphics[width=0.80\linewidth]{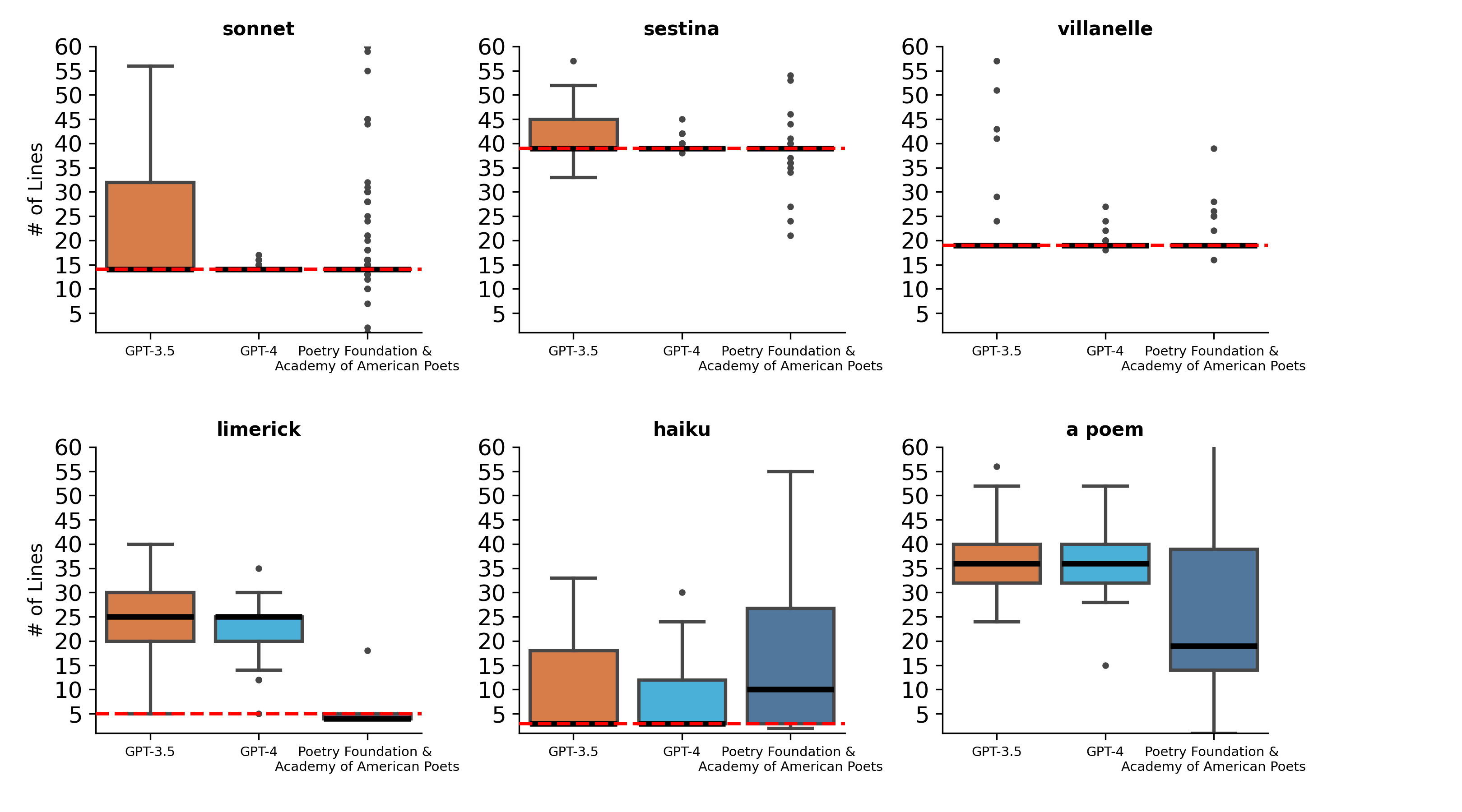}
    \caption{These boxplots represent the distribution of line lengths for poems with conventionally fixed lengths produced by \textbf{\textcolor{GPT3}{GPT-3.5}}, \textbf{\textcolor{GPT4}{GPT-4}}, and authors from the \textbf{\textcolor{PF}{Poetry Foundation and the Academy of American Poets}}. The GPT models were also prompted with the generic style of ``a poem''; to provide a comparison for the human poems, we include an aggregation of all poems from the sample. The boxes show the ``interquartile range'' (25\% quartile-75\% quartile) with a thicker line indicating the median average; the whiskers extend beyond the boxes by  1.5 times the IQR; the outliers are values that fall beyond the whiskers. The dotted red line indicates the expected number of lines for each form, e.g., a sonnet typically has 14 lines.}
    \label{fig:fixed-avg-line}
\end{figure}

\begin{figure}[p]
    \centering
\includegraphics[width=0.87\linewidth]{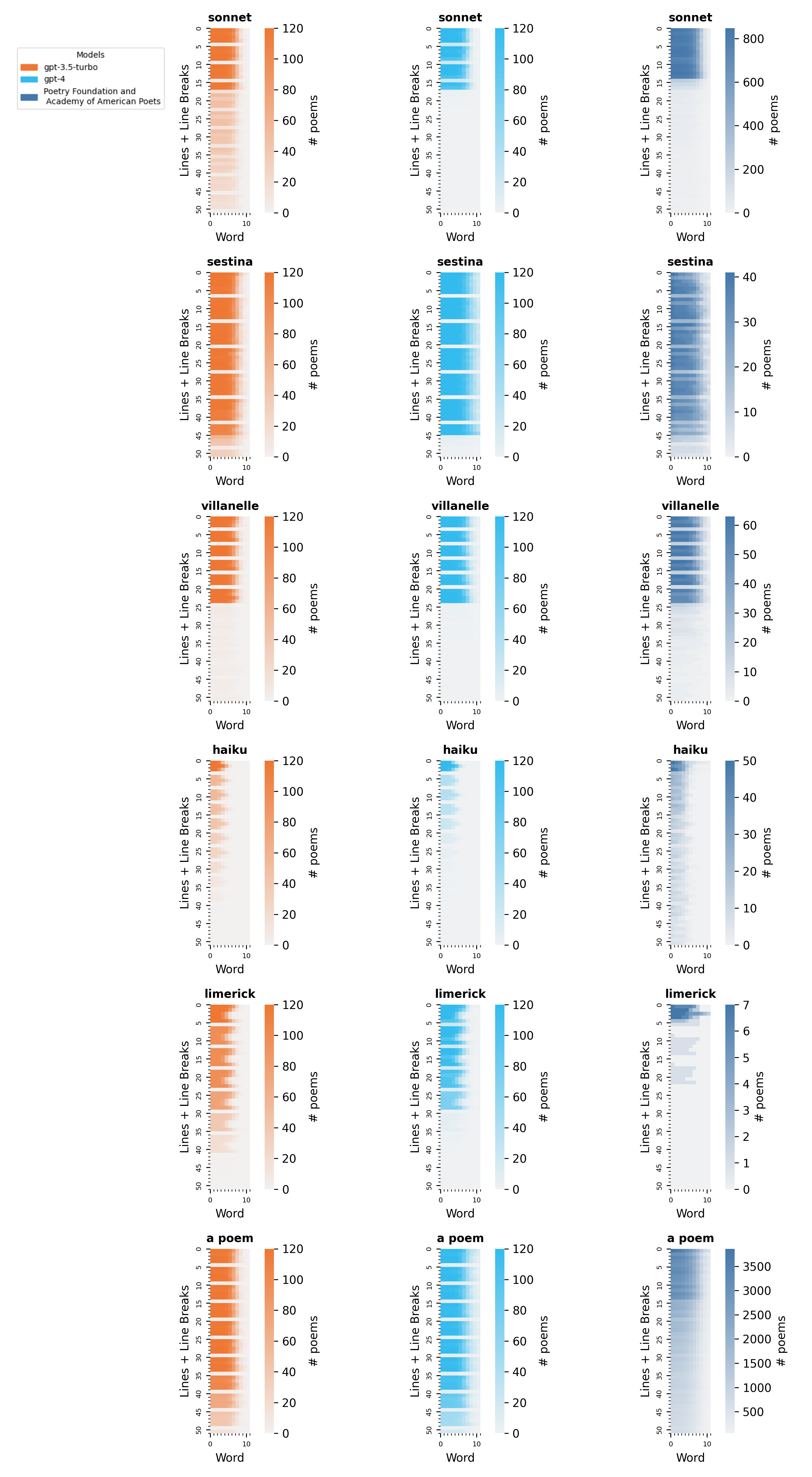}
    \caption{These heatmaps represent the distribution of words, lines, and line breaks for \textit{fixed form }poems by \textbf{\textcolor{GPT3}{GPT-3.5}}, \textbf{\textcolor{GPT4}{GPT-4}}, and authors from the \textbf{\textcolor{PF}{Poetry Foundation and the Academy of American Poets}}. Darker squares represent a higher concentration of words and lines in specific positions across the poems; lighter squares represent a higher concentration of white space and line breaks. The GPT models are also prompted with the generic style of ``a poem''; to provide a comparison for the human poems, we include an aggregation of all poems from the sample.}
    \label{fig:heatmap}
\end{figure}

We measure the number of lines and the number and kind of stanzas across all the poems by parsing line breaks. We visualize these distributions as boxplots (Figure \ref{fig:fixed-avg-line}) and heatmaps (Figure \ref{fig:heatmap}), revealing the most common lengths and shapes of the poems across styles and forms. These results show that when we prompt the models to generate poems in forms with typically fixed lengths---such as \textit{sonnets} (14 lines)---they largely adhere to this convention, with notable improvement in GPT-4. While GPT-3.5 and GPT-4 both generate sonnets with a median average length of 14 lines, Figures \ref{fig:fixed-avg-line} and \ref{fig:heatmap} show that there is much more variability in GPT-3.5. As displayed in the boxplot, the upper 75\% quartile extends to 32 lines, and the range extends to 55 lines.  
By contrast, the entire range of GPT-4 sonnets (minus outliers) falls at exactly 14 lines. Line lengths for sestinas (typically 39 lines) and villanelles (typically 19 lines) follow a similar pattern. The median lengths are appropriately 39 and 19 lines for both models, but GPT-4 demonstrates much more consistency, displaying a smaller interquartile range and spread of outliers (the same consistency is also displayed in the heatmap in Figure \ref{fig:heatmap}).

Interestingly, for these three forms, GPT-4 hues closer to ``conventional'' lengths than our sample of poems from the Poetry Foundation and the Academy of American Poets.
In a small percentage of these human poems, we find that the longer lengths come from explicit or implicit play with or defiance of the forms. For example, Bino A. Realuyo concludes his 15-line poem, \href{https://poets.org/poem/eulers-equation}{``Euler's Equation,''} with the line: ``a rebellion, the fifteenth line of a sonnet.'' More often, in these longer poems, authors include a given form in multiples, such as Algernon Charles Swinburne's ``double sestina'' (12 stanzas of 12 lines each) in \href{https://www.poetryfoundation.org/poems/45281/the-complaint-of-lisa}{``The Complaint of Lisa.''} 

An obvious aberration for the GPT models is their atypically long limerick style. Where a traditional limerick is usually about 5 lines long, the median length for both GPT models is 25 lines. Upon closer inspection, it is clear that both models frequently bundle multiple, appropriately-lengthed limericks together. The heatmap in Figure \ref{fig:heatmap} shows that the models often produce several limericks in a row. This is also the case for GPT-3.5's atypically long sonnets, which are usually multiple sonnets packed into one. While these multiples resemble the long poems that we observe in our human poetry sample, we think this tendency more likely suggests that, in certain cases, the GPT models know \textit{how} to produce a particular kind of poem but don't know \textit{when} to stop.

\begin{table*}[h]
    \centering
    \scriptsize
    \caption{\textbf{Quatrains.} Percentage of poems with at least one quatrain and percentage of quatrains of all stanzas.}
    \begin{tabular}{p{6cm}|p{4cm}|p{4cm}|}
    \toprule
    \textbf{Source}& \textbf{Poems with Quatrain}& \textbf{Stanzas with Quatrain}\\
    \midrule
    Poetry Foundation and Academy of American Poets & 713   / 3,874 poems  (\textbf{18.4\%})& 3,014  / 18,052 stanzas 
 (\textbf{16.7\%})\\
    GPT-3.5 Turbo & 2,027 / 2,880 poems (\textbf{70.4\%})& 16,089 / 24,093 stanzas (\textbf{66.8\%})\\
    GPT-4 & 1,824 / 2,880 poems (\textbf{63.3\%})& 13,303 / 22,305 stanzas (\textbf{59.6\%})\\
    \bottomrule
    \end{tabular}
    
    \label{table:quatrain-distribution}
\end{table*}

Aside from limericks, the GPT models can broadly produce poems of appropriately diverse lengths for a range of fixed forms, but they demonstrate a penchant for producing poems of an almost ``default'' size when left to their own devices. For both models, the median average length for a generic ``poem'' is 36 lines (see Figure \ref{fig:fixed-avg-line}), and the overall median length across all styles is 32 lines. 

\begin{figure}[h]
    \centering
    \includegraphics[width=.75\linewidth]{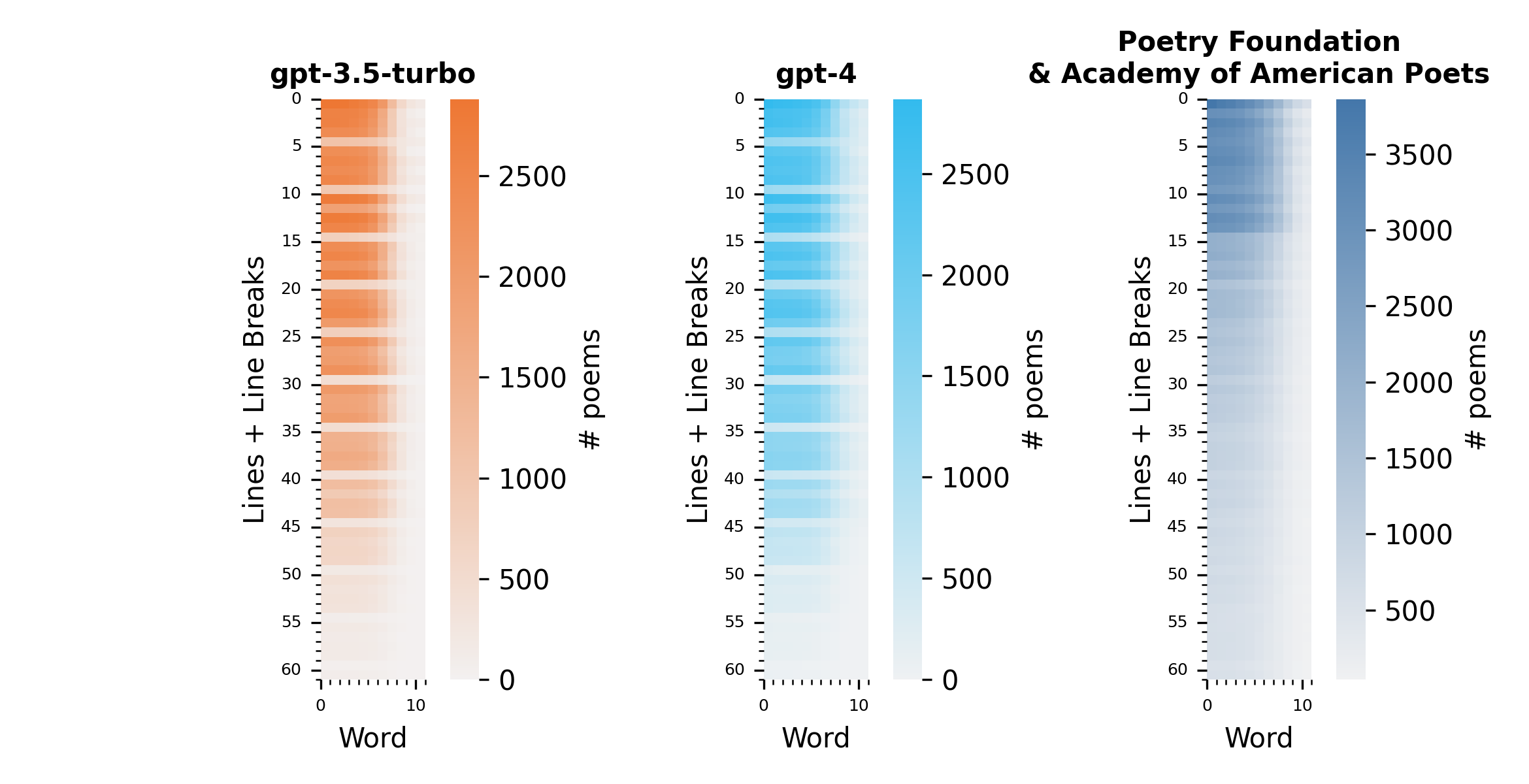}
    \caption{These heatmaps represent the distribution of words, lines, and line breaks for all poems by \textbf{\textcolor{GPT3}{GPT-3.5}}, \textbf{\textcolor{GPT4}{GPT-4}}, and authors from the \textbf{\textcolor{PF}{Poetry Foundation and the Academy of American Poets}}. Darker squares represent a higher concentration of words and lines in specific positions across the poems; lighter squares represent a higher concentration of white space and line breaks. The unusual dominance of quatrains (line breaks after 4 consecutive lines) is evident in the GPT-generated poems.}
    \label{fig:box-plot-styles}
\end{figure}

Another striking feature of the GPT poems is the dominance of 4-line stanzas, or \textit{quatrains}. We find that while just 16.7\% of the human-authored stanzas are quatrains, a whopping 66.8\% of all GPT-3.5 stanzas and 59.6\% of all GPT-4 stanzas are quatrains (Table \ref{table:quatrain-distribution}). The heatmaps in Figure \ref{fig:box-plot-styles} visually demonstrate how common quatrains are across the GPT-generated poems, showing clear line breaks in regular 4-line intervals, with no such regularity evident in the human poems.

\subsection{Collective Perspective}

\begin{figure*}[h]
    \centering
    \includegraphics[width=1\linewidth]{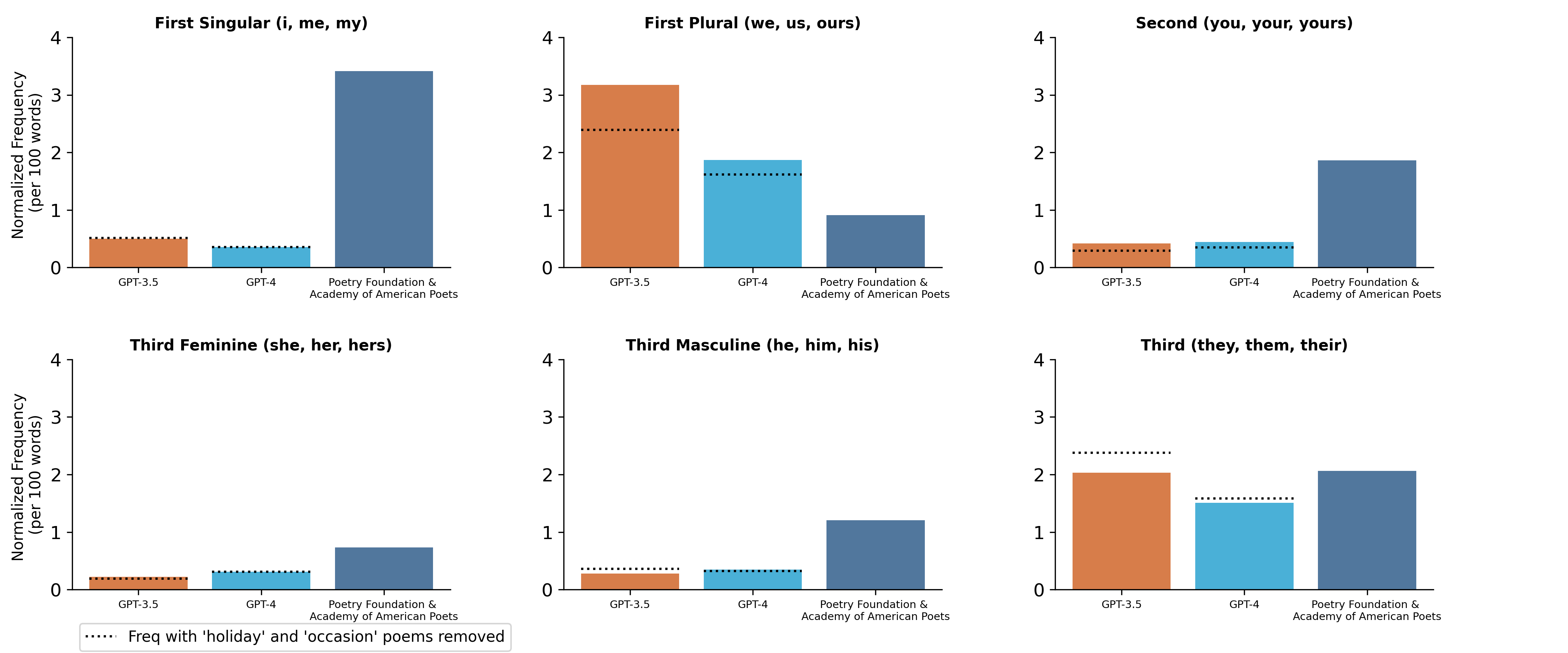}
    \caption{The normalized frequency of pronouns used in poems by \textbf{\textcolor{GPT3}{GPT-3.5}}, \textbf{\textcolor{GPT4}{GPT-4}}, and authors from the \textbf{\textcolor{PF}{Poetry Foundation and the Academy of American Poets}}, expressed per 100 words. The dotted line indicates normalized frequency in the GPT poems with the ``holiday'' and ``occasion'' poems removed (showing that first-person plural in the GPT-generated poems decreases slightly, and third-person increases slightly).}
    \label{fig:avg-pronouns}
\end{figure*}

We measure the normalized frequency of pronouns (Table \ref{tab:pronouns}) in each corpus, expressed per 100 words. We find that poems produced by GPT-3.5 and GPT-4 tend to use more first-person plural pronouns (``we,'' ``us,'' ``our'')  and fewer first-person singular pronouns (``i'', ``me,'' ``myself'') than poems written by humans (see Figure \ref{fig:avg-pronouns}). For example, GPT-4 produced the following limerick about Memorial Day in response to our \textit{figurative prompt} (which specifies not to include the style or subject words in the poem):

\begin{quote} 

\textit{In May \textbf{we} stand strong, hearts ablaze,\newline 
For those who've seen war's smoky haze.\newline 
\textbf{We} honor the brave,\newline 
Who life for \textbf{us} gave,\newline 
In silence, \textbf{we} give them \textbf{our} praise.}
\end{quote}

This limerick continues on for 20 more lines in 5-line, rhyming (AABBA) stanzas, consistent with the results presented in Section \ref{length}.

Because there are a large number of ``holiday'' and ``occasion'' subject prompts like ``Memorial Day,'' which perhaps encourage meditation on collective experiences, we also show normalized frequency for the GPT-generated poems with these subjects removed (see the dotted lines in Figure \ref{fig:avg-pronouns}). Without these subjects, the normalized frequency for the first-person plural decreases slightly, and it increases slightly for the third-person. But the curious dominance of the first-person plural is still present. We think this pattern may reflect the models' pre-programmed attitudes toward inclusivity, as well as its obvious lack of first-person singular experiences, but more work is needed to explore this trend further.

\begin{table*}[h]
    \caption{Pronouns by Category}
  \begin{tabular}{ll}
    \toprule
    Category & Pronouns \\
    \midrule
    First Singular & i, me, my, mine, myself \\
    First Plural & we, us, our, ours, ourselves \\
    Second & you, your, yours, yourself, yourselves, thou, thee, thy, thine, thyself \\
    Third Feminine & she, her, hers, herself \\
    Third Masculine & he, his, him, himself \\
    Third & they, them, their, theirs, themself, themselves, it, its, itself \\
    \bottomrule
  \end{tabular}
  \label{tab:pronouns}
\end{table*}


\subsection{Most Distinctive Words}

\begin{figure*}[h]
    \centering
    \begin{subfigure}[b]{0.48\linewidth}
        \centering
        \includegraphics[width=\linewidth]{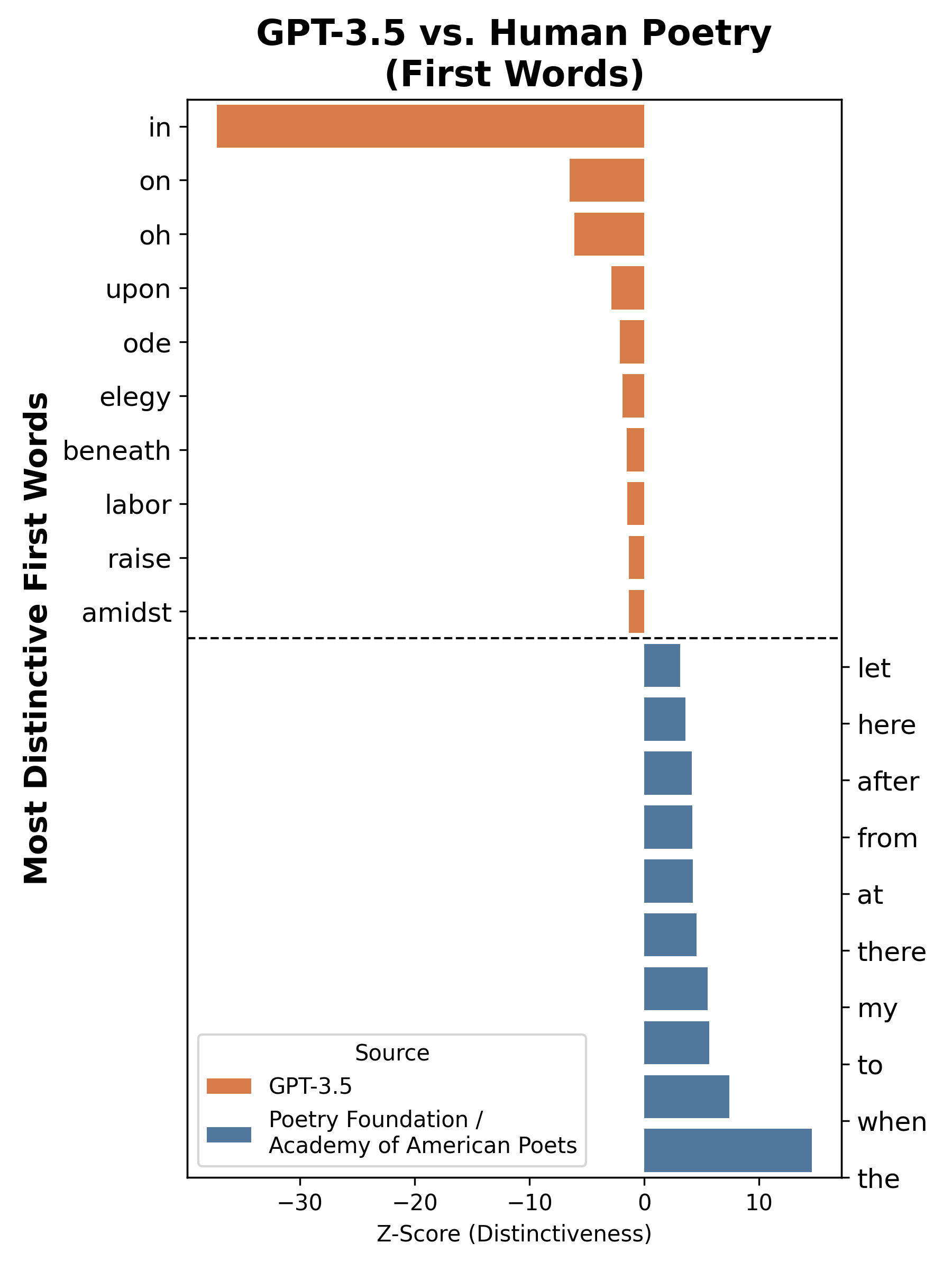}
        
        \label{fig:top-words-gpt35}
    \end{subfigure}
    \hfill
    \begin{subfigure}[b]{0.48\linewidth}
        \centering
        \includegraphics[width=\linewidth]{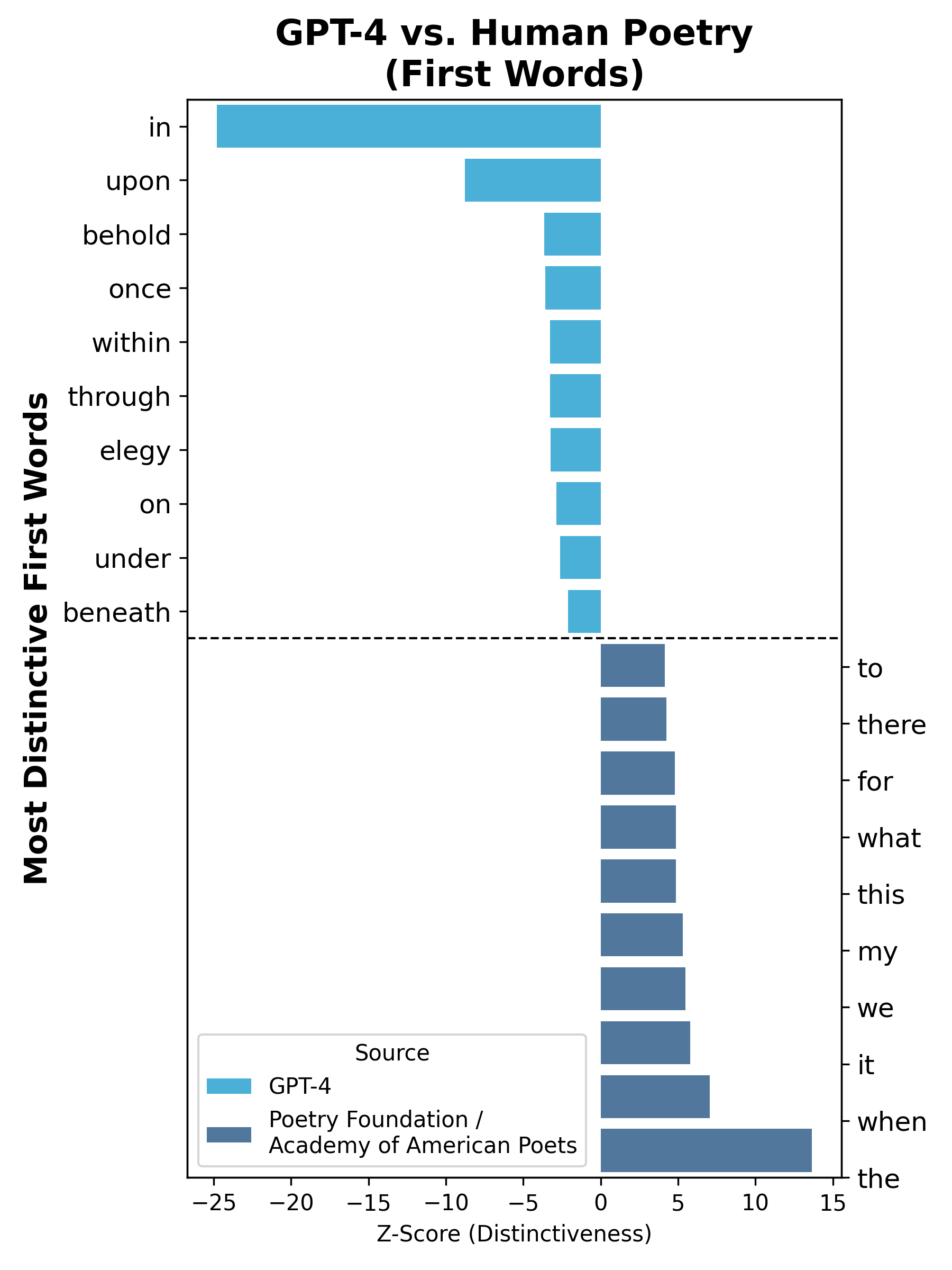}
       
        \label{fig:first-words-gpt4}

    \end{subfigure}
    \caption{The 10 most distinctive first words in poems produced by \textbf{\textcolor{GPT3}{GPT-3.5}}, \textbf{\textcolor{GPT4}{GPT-4}}, or the \textbf{\textcolor{PF}{Poetry Foundation and the Academy of American Poets}}. To identify these words, we use \citet{monroe_fightin_2017}'s algorithm for  comparing language use across text corpora. Stopwords are not removed.}
    \label{fig:side-by-side-first}
\end{figure*}

\begin{figure*}[h]
    \centering
    \begin{subfigure}[b]{0.48\linewidth}
        \centering
        \includegraphics[width=\linewidth]{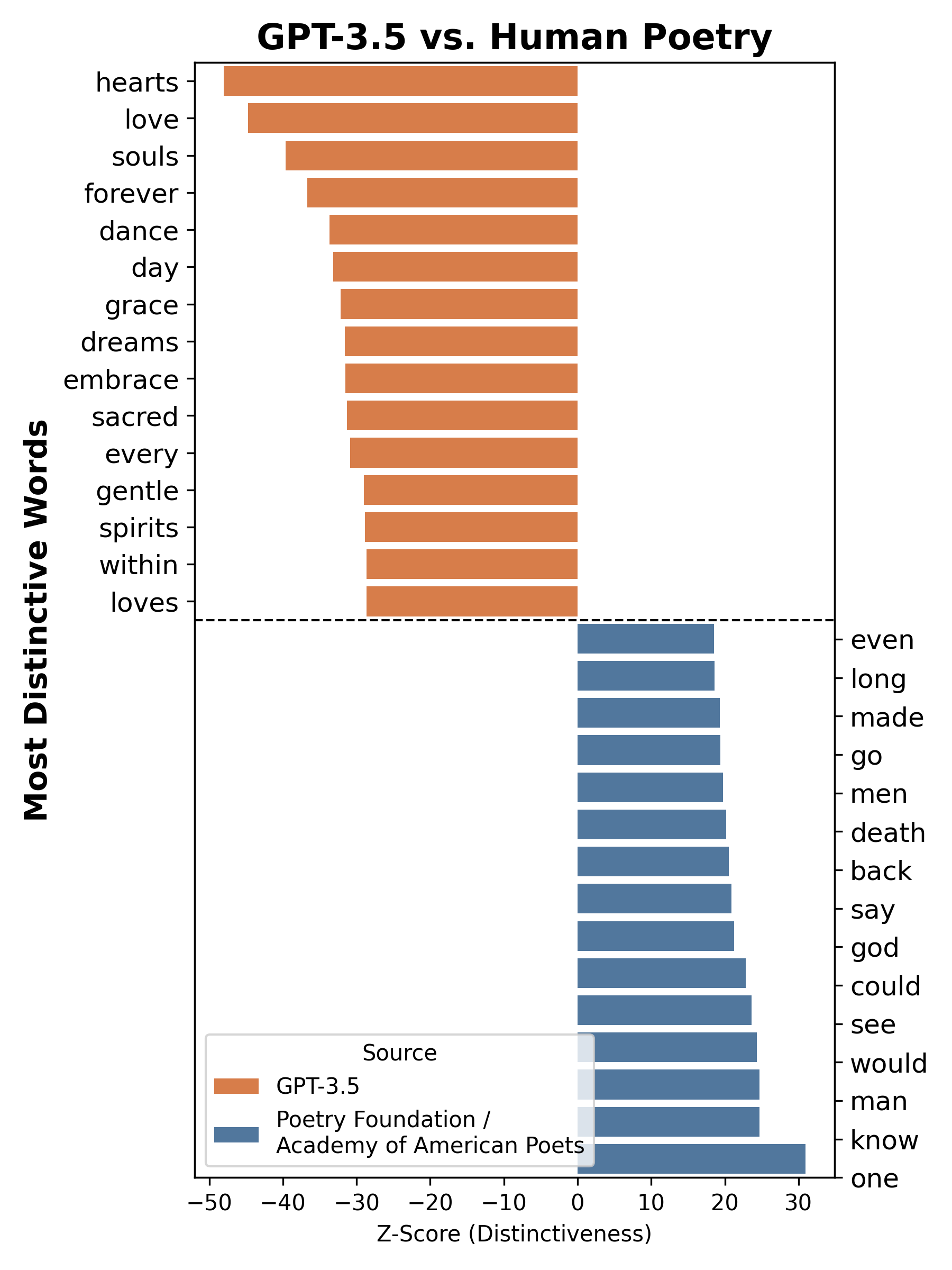}
        \label{fig:top-words-gpt35}
        
    \end{subfigure}
    \hfill
    \begin{subfigure}[b]{0.48\linewidth}
        \centering
        \includegraphics[width=\linewidth]{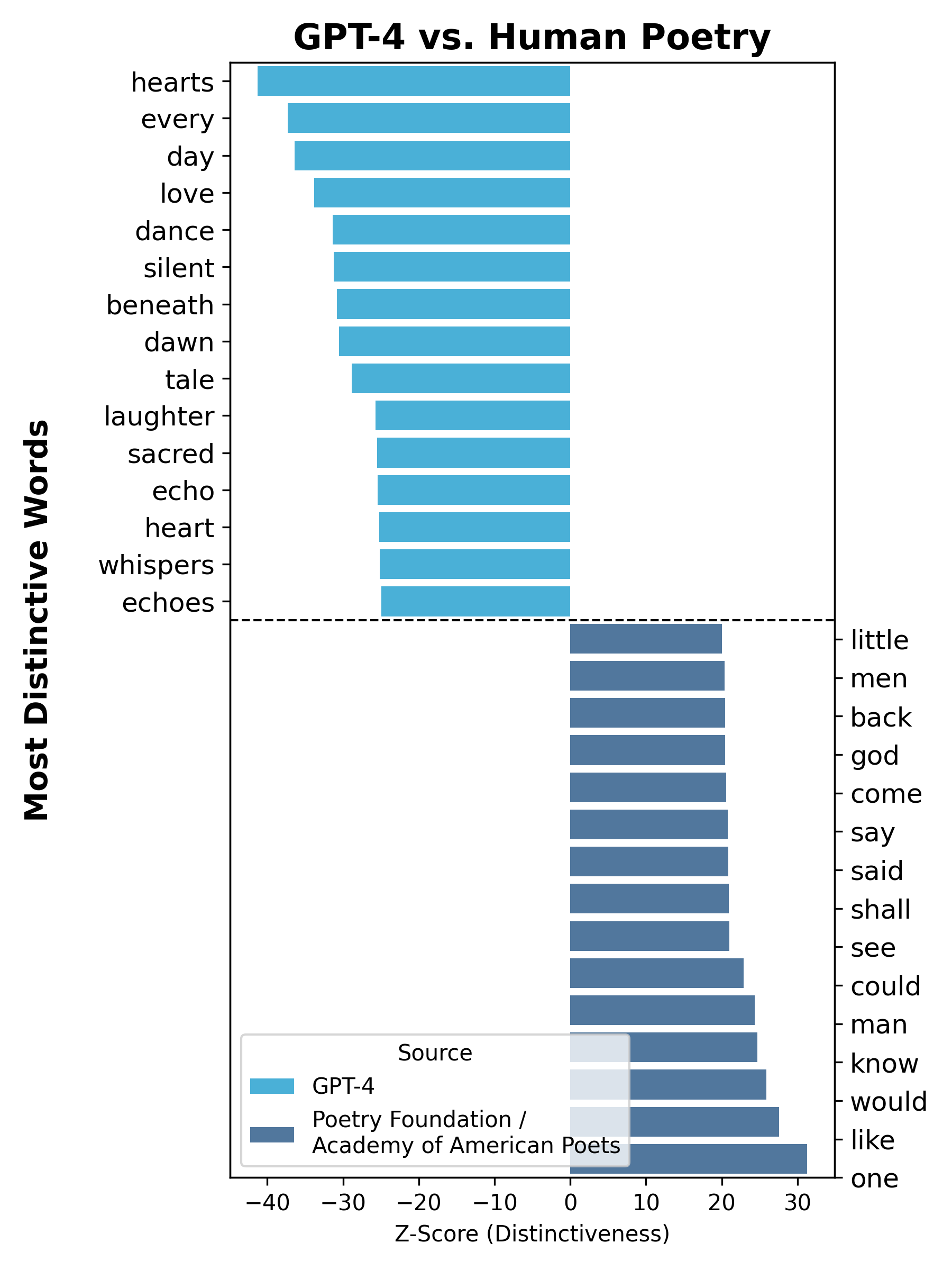}
        \label{fig:top-words-gpt4}
        
    \end{subfigure}
    \caption{The 15 most distinctive words in poems produced by \textbf{\textcolor{GPT3}{GPT-3.5}} and \textbf{\textcolor{GPT4}{GPT-4}} vs. poems found in the \textbf{\textcolor{PF}{Poetry Foundation and the Academy of American Poets}}. To identify these words, we use \citet{monroe_fightin_2017}'s algorithm for  comparing language use across text corpora. In this case, stopwords are removed.}
    \label{fig:side-by-side}
\end{figure*}

We also analyze the most distinctive opening words and overall words across the poems using \citet{monroe_fightin_2017}'s ``fightin' words'' algorithm, which uses weighted log-odds ratios with an informative Dirichlet prior. This method is designed to robustly compare word usage across unevenly distributed text corpora. We specifically use an implementation by \citet{hessel_jmhesselfightingwords_2024}\footnote{\href{ttps://github.com/jmhessel/FightingWords}{ttps://github.com/jmhessel/FightingWords}} and restrict the vocabulary to words that appear in a minimum of 10 poems.
We remove stopwords for our overall word analysis but not for our first word analysis. In Figures \ref{fig:side-by-side-first} and \ref{fig:side-by-side}, we display words with the highest Z-scores for each category, representing the most distinctive words.

The most distinctive opening word in both the GPT-3.5- and GPT-4 generated poems is ``In'' (see Figure \ref{fig:side-by-side-first}). This preposition is included across a wide range of poems and contexts, such as:

\begin{quote}
\textit{
\textbf{In} autumn's blaze of golden hue... (GPT-3.5 --- pantoum, Thanksgiving)\newline
\textbf{In} the darkest days, a flicker of light...(GPT-3.5 --- ars poetica, Hanukkah)\newline
\textbf{In} the girth of world-kaleidoscope, we are birthed into living, (GPT-4 --- free verse, living) \newline}
\end{quote}

\noindent The next most distinctive first word in GPT-4 poems is ``Upon,'' which also seems to be a frequent way for the model to initiate iambic meter:

\begin{quote}
\textit{
\textbf{Upon} a stage where shadows nightly reign... (GPT-4 --- sonnet, Halloween)\newline
\textbf{Upon} this day, we sing the laborer's song,... (GPT-4 --- sonnet, Labor Day)\newline
\textbf{Upon} the chill of winter's breath descends,... (GPT-4 --- blank verse, Hanukkah) \newline}
\end{quote}

\noindent The word ``upon'' is an iamb (the basic unit of iambic meter), meaning it consists of an unstressed syllable followed by a stressed syllable. This is also the case for other distinctive GPT-4 first words, such as ``beneath,'' ``behold,'' and ``within.'' First words in poems from the Poetry Foundation and Academy of American Poets show no such distinctive patterns, mostly consisting of articles and pronouns.


Overall, the distinguishing vocabulary for the GPT models consists of words associated with love (``heart,'' ``love,'' ``souls,''), words that rhyme (``grace'', ``embrace''), and words that are acoustic (``echo,'' ``whisper''). For GPT-3.5, words like ``embrace,'' ``grace,'' ``dance,'' and ``dreams'' are touchstones. At least one of these words shows up in 87\% of the GPT-3.5 poems. For GPT-4, either ``echo'' or ``whisper'' shows up in 75\% of the poems. For example, both words appear in this \textit{dramatic monologue} about the \textit{arts \& sciences}:

\begin{quote}
\textit{Look upon me, ageless I stand, the crossing of arts and sciences,\newline 
In the \textbf{echoing} hallways of knowledge, beneath glimmering frescoed edifices.\newline
I am the \textbf{whisperer} in marbled alcoves, the scribe of thinkers’ existence,\newline
Caught in endless dialogue, between creativity and discipline's persistence.\\
-GPT-4 (dramatic monologue, arts \& sciences)\newline}
\end{quote}

\subsection{Prosody Analysis (Rhyme \& Meter)}

``Prosody'' refers to patterns of sound in poetry, encompassing rhyme and meter. 
Analyzing prosody across a large corpus poses challenges even in human-authored poems because it relies on the pronunciation of particular words in relation to each other. 
The text of a poem does not provide direct access to its prosody because the same word may have different pronunciations in different forms of English, and even with the same pronunciation a word might be stressed or unstressed depending on its context. For example, in Alfred, Lord Tennyson's poem \href{https://www.poetryfoundation.org/poems/45392/ulysses}{``Ulysses,''} the word ``I'' is unstressed at the beginning of a line and stressed toward the end: ``I cannot rest from travel: I will drink.'' 
Analyzing prosody across GPT-generated poems poses additional challenges because unlike human-authored poems, which are often either clearly free verse or clearly aimed at embodying a particular metrical pattern, the meter of GPT outputs can be less precise and harder to define with a single metrical label. 
 


\begin{table*}[h]
    \centering
    \scriptsize
    \caption{\textbf{Rhyme Usage.} Percentage of poems with rhyme and average percentage of rhymed lines, based on quantitative analysis with the CMU Pronouncing Dictionary. Rhymed lines include AA, ABAB, ABBA, and ABCB rhymes.}
    \begin{tabular}{p{6cm}|p{4cm}|p{4cm}|}
    \toprule
    \textbf{Source} & \textbf{Poems with at least One Rhyme} & \textbf{Avg. Percent Rhymed Lines} \\
    \midrule
    Poetry Foundation and Academy of American Poets & 2,518 / 3,874 poems (\textbf{65.0\%}) & 29.45\% \\
    GPT-3.5 Turbo & 2,599 / 2,880 poems (\textbf{90.2\%}) & 63.87\% \\
    GPT-4 & 2,578 / 2,880 poems (\textbf{89.5\%}) & 65.20\% \\
    \bottomrule
    \end{tabular}
    
    \label{table:rhyme-distribution}
\end{table*}

To measure prosody, we thus conduct both a quantitative \textit{and} qualitative analysis. 
For our manual analysis, we take a random sample of poems in each form and, drawing on our domain expertise, hand-annotate various prosodic elements where they are discernible, including dominant meter, line-length (in terms of poetic feet---tetrameter, pentameter, etc.), rhyme scheme, and stanza patterns. We evaluate 144 poems produced by GPT-3.5 (6 in each form) and 144 poems produced by GPT-4 (6 in each form) for just over 5\% of the GPT-generated corpus. We also analyze 138 poems from the human-authored corpus (6 poems in 23 forms), making up just over 3.7\% of the human-authored corpus.

Over 80\% of the GPT-generated poems in our random sample contain patterns of end rhyme, as compared with around 50\% of the human-authored poems. Over 60\% of the GPT-authored poems had a dominant iambic meter, compared to just under 40\% of poems from the human corpus. When we break these results down between the GPT-3.5 and GPT 4 models, it appears that the dominance of iambic meter is lessening somewhat in the newer model. Only around 53\% of GPT-4-authored poems had a dominant iambic meter compared to almost 74\% of GPT-3.5-authored poems. Rather than indicating a shift in the model's default tendencies in relation to poetry, we think this change may reflect GPT-4's increased ability to \textit{not} produce iambic meter when it is prompted to produce poems in forms that do not traditionally include regular meter---for example haiku, prose poetry, or free verse.

\begin{figure}[h]
    \centering
    \includegraphics[width=.95\linewidth]{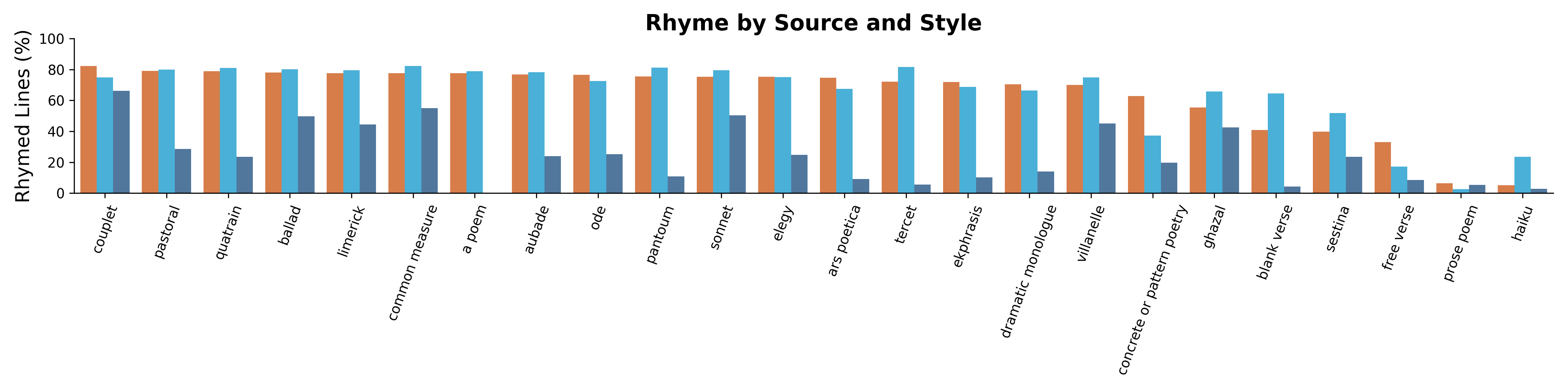}
    \caption{These bar plots show percentage of lines rhymed in poems by \textbf{\textcolor{GPT3}{GPT-3.5}}, \textbf{\textcolor{GPT4}{GPT-4}}, and authors from the \textbf{\textcolor{PF}{Poetry Foundation and the Academy of American Poets}}. Rhymes were calculated with the CMU Pronouncing Dictionary and include AA, ABAB, ABBA, and ABCB rhymes. Styles are sorted from most to least rhyme based on poems by the GPT-3.5 model.}
    \label{fig:rhyme-style}
\end{figure}

We also conduct a quantitative analysis of rhyme across all poems using \citet{parrish_aparrishpronouncingpy_2024}'s Python implementation of the CMU Pronouncing Dictionary.\footnote{\href{https://github.com/aparrish/pronouncingpy}{https://github.com/aparrish/pronouncingpy}} Using this dictionary, we measure how many lines end with AA, ABAB, ABBA, or ABCB rhymes. Around 90\% of the poems generated by both the GPT-3.5 and GPT-4 models contained at least one end rhyme, compared to around 65\% of our human-authored corpus. Moreover, for certain forms like \textit{sonnets} and \textit{ballads}, the models \textit{always} produced poems with at least one end rhyme. This was also the case for unfixed forms like \textit{aubade} or \textit{pastoral}, which do not typically require specific patterns of rhyme. The percentage of rhyming lines for all styles and sources is displayed in Figure \ref{fig:rhyme-style}.


Taken together, our analyses suggest that there is a sort of default poetic mode in GPT models, which favors quatrains, iambic meter, and end rhyme. The models can be prompted to produce writing in other styles, but sometimes the persistent iambic/quatrain/end rhyme style still breaks through. For example, although prompted to produce a \textit{social commentary}-related \textit{limerick}, which would typically be a five-line stanza with anapestic meter and AABBA rhyme, the GPT-4 model produced the following five quatrains with iambic meter and AABB rhyme:

\begin{quote}
\begin{multicols}{2}
\small{\textit{A world viewed through a smartphone's \textbf{lens},\\
Each day a stream of trends \textbf{begins},\\ 
Tales of fortune, tales of\textbf{ woes},\\ 
Each one judged as the wind \textbf{blows}.} 

\textit{In the glow of public opinions' \textbf{glare},\\
Endless chatter fills the \textbf{air}.\\ 
Russian bots to kitty \textbf{memes},\\ 
Nothing's truly as it \textbf{seems}.} 

\textit{Their thoughts they tweet, in speech so \textbf{free},\\ 
Behind the screens, who can they \textbf{be?}\\ 
Worldly wisdom, or just \textbf{noise?}\\ 
Undiscovered truth, or \textbf{toys?}} 

\textit{As they debate who's wrong or \textbf{right,}\\ 
Their fingers dance in virtual \textbf{fight.}\\ 
From east to west, the judgments \textbf{fly,}\\ 
Sometimes truth, oftentimes \textbf{lie.}}

\textit{We see the world in constant \textbf{chime,}\\ 
A universe of views \textbf{online,}\\
In every tweet and post we \textbf{glean,}\\ 
A reflection of a world \textbf{unseen.}}}
\end{multicols}
\end{quote}

While this poem is a funny and topical take on a social commentary (``Russian bots to kitty memes / Nothing's truly as it seems''), it also highlights some of the model's limitations. Even when instructed to produce more specific or varied forms, the model’s default poetic mode--- quatrains, iambic meter, and end rhyme---often resurfaces, favoring structured patterns over more diverse styles.


\section{Discussion}\label{discussion}

By showing that GPT-generated poetry contains clear stylistic quirks and characteristics, we build upon emerging research on the style of LLM-generated texts \cite{cheng_marked_2023}. For example, after analyzing millions of biomedical article abstracts, \citet{kobak_delving_2024} show that ``hundreds of words have abruptly increased their frequency after ChatGPT became available.'' They reveal particular spikes in the use of ``style-affecting verbs and adjectives that ChatGPT-like LLMs prefer,'' such as ``delve,'' ``significant,'' and ``crucial.'' Although \citet{kobak_delving_2024} are primarily interested in LLMs' impact on academic research, their findings contribute to growing knowledge about LLM style in specific genres. Our findings that the GPT models are more homogeneous and standardized than human poets also aligns with similar work that has shown LLM poetry to be ``underdiverse'' and restricted, especially with features like rhyme \cite{chen_evaluating_2024, martinHistoricalProsodyMechanical2024}. We believe there are more exciting opportunities for digital humanities scholars and language experts to continue studying these more cultural and artistic dimensions of LLM-generated texts.

The results of our poetry prompting experiments highlight both advancements and notable limitations with GPT-generated poetry. The models' ability to produce poems of appropriate lengths for a wide variety of forms and styles---without any fine-tuning---marks a significant development in automatic poetry generation, especially since the models manage to do so while incorporating rhyme and meter and maintaining general clarity. Yet, overall, the models also exhibit far less variation, diversity, and creativity than the human-authored poems.
However, we want to flag that \textit{prompting}---what the user asks the model to generate and how that ask is constructed---plays a major role in shaping the poetry that the models produce. In related experiments, when we prompted the models with specific author names, our results seemed to shift and become more complex. 
In this specific study, our goal was not to produce the most creative and interesting poetry possible, but rather to understand the broad contours of the models and their outputs. 
If we wanted to produce more interesting poetry, we would likely use different prompts. 

\section{Conclusion}

We prompt the GPT-3.5 and GPT-4 models to generate English-language poems in 24 different poetic forms or styles, about 40 different subjects, and in response to 3 different templates. We compare these GPT-generated poems to a sample of poems from the Poetry Foundation and the Academy of American Poets, showing that the GPT models are much more formulaic and constrained than the human-authored poetry. We argue that the GPT models have a ``default'' poetic mode, characterized by quatrains with rhymed lines in iambic meter; first-person plural perspectives; and the repetition of words like ``heart,'' ``embrace,'' ``echoes,'' and ``whispers.'' This default mode sometimes breaks through even when otherwise prompted. We share the code that we used to conduct this analysis,\footnote{\href{https://github.com/melaniewalsh/chatgpt\_poetry}{https://github.com/melaniewalsh/chatgpt\_poetry}} and we share all the public domain human-authored poems and form/style annotations from prior work.\footnote{\href{https://github.com/maria-antoniak/poetry-eval}{https://github.com/maria-antoniak/poetry-eval}} We plan to share our ChatGPT poetry corpus at a later stage of the project. In future work, we plan to explore a wider range of prompts (potentially including author names) and models, and to study the poems more closely with traditional literary studies approaches.  

\begin{acknowledgments}
  We would like to thank the reviewers for their thoughtful and very helpful feedback. We would also like to thank Zoe LeBlanc, John Ladd, Matt Lavin, and Gabi Kirilloff for early conversations and feedback about this work. This research was partly supported by the NEH-funded "AI for Humanists" project and by an "AI, Creativity, and Humanities" collaboration grant from the Simpson Center for the Humanities at the University of Washington.
\end{acknowledgments}

\clearpage

\bibliography{main}




\end{document}